\documentclass{article}

\usepackage{multirow}
\usepackage{arxiv}

\usepackage[utf8]{inputenc} % allow utf-8 input
\usepackage[T1]{fontenc}    % use 8-bit T1 fonts
\usepackage{hyperref}       % hyperlinks
\usepackage{url}            % simple URL typesetting
\usepackage{booktabs}       % professional-quality tables
\usepackage{amsfonts}       % blackboard math symbols
\usepackage{nicefrac}       % compact symbols for 1/2, etc.
\usepackage{microtype}      % microtypography
\usepackage{lipsum}		% Can be removed after putting your text content
\usepackage{graphicx}
\usepackage{natbib}
\usepackage{doi}
\usepackage{latexsym}
\usepackage{longtable}
\usepackage{algpseudocode}

\title{Opinion Mining using Population-tuned\\ Generative Language Models}

\author{ \href{https://orcid.org/0000-0002-3926-6653}{\includegraphics[scale=0.06]{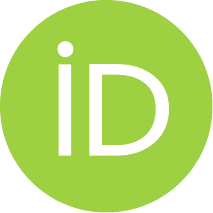}\hspace{1mm}Allmin Susaiyah} \\
	Dept. of Mathematics and Computer Science\\
	Eindhoven University of Technology\\
	Netherlands \\
	\texttt{a.p.s.susaiyah@tue.nl} \\
\And
\href{https://orcid.org/0000-0002-3261-2920}{\includegraphics[scale=0.06]{orcid.pdf}\hspace{1mm}Abhinay Pandya} \\
Language Team\\
Philips Research, Eindhoven\\
The Netherlands\\
\texttt{abhinay.pandya@philips.com} \\
	 \AND
\href{https://orcid.org/0000-0002-2966-3305}{\includegraphics[scale=0.06]{orcid.pdf}\hspace{1mm}Aki H{\"a}rm{\"a}} \\
	 Advanced Computing Sciences \\
	 Maastricht University, The Netherlands \\
	 \texttt{aki.harma@maastrichtuniversity.nl} \\
}

\renewcommand{\shorttitle}{\textit{arXiv} Template}

\hypersetup{
pdftitle={Opinion Mining using Population-tuned Generative Language Models},
pdfsubject={q-cs.CL},
pdfauthor={Allmin Susaiyah},
pdfkeywords={Opinion Mining, Large Language Models, Insight Generation},
}

\begin{document}
\maketitle

\begin{abstract}
We present a novel method for mining opinions from text collections using generative language models trained on data collected from different populations. We describe the basic definitions, methodology and a generic algorithm for opinion insight mining. We demonstrate the performance of our method in an experiment where a pre-trained generative model is fine-tuned using specifically tailored content with unnatural and fully annotated opinions. We show that our approach can learn and transfer the opinions to the semantic classes while maintaining the proportion of polarisation. Finally, we demonstrate the usage of an insight mining system to scale up the discovery of opinion insights from a real text corpus.
\end{abstract}

% keywords can be removed

\section{Introduction}

% #todo: mapping list

In recent years, transformer-based generative pre-trained language models such as the GPT2 \cite{gpt2}, GPT3\cite{brown2020language}, GPT-Neo \cite{gao2020pile,kashyap2022gpt} and OPT \cite{Zhang2022OPTOP} have gained popularity because of their ability to perform well in a variety of NLP tasks such as machine translation and question answering. 
 The paper introducing the famous GPT3 generative language model \cite{brown2020language} devoted four pages to a detailed analysis of various biases in gender, race, and religion in the text the model generates. 
 Language evolved in early hominins as a tool for conversation that "expresses our highest aspirations, our basest thoughts, and our philosophies of life" \cite{everett2017language}. Those are inseparable parts of communication and a language model likely learns those expressions from any collection of natural language content. Conversely, if we discover those from the output of the model, we could learn about the thoughts of the population that produced the training content.

 In data-to-text insight generation tasks, see, e.g., \cite{Reiterdata2text,sumtime,harma2016probabilistic} an {\it insight} is often defined as a categorical statement about a measure in two contexts \cite{susaiyah2020towards}, for example, {\bf apples are bigger than pears}. Let us define an {\it opinion insight} as a thought of a population about a certain entity that takes the form of such an insight. In the absence of targeted surveys and tabular results of such surveys, it is possible to find opinions like the one above using textual corpora. Such opinions could be stratified by selecting discourse corpora from various subgroups; for example, the classical Greeks or left-handed people, etc. The sentiment polarity evaluation of such text segments towards the entities of interest can then provide an indication of the opinion of the target population towards the selected entities. 
 Traditionally, such opinion insights have been based on questionnaire study data, for example, asking left- and right-handed people about their opinions about the sizes of different fruits. Questionnaire studies are expensive, time-consuming, and require careful design of the questions we are interested in in advance. 
 
 The basic idea of the current paper is to replace the questionnaire studies with generative language models trained on the target population. Opinion insights can be derived by analyzing the outputs from a generative language model (GLM) such as the GPT2 \cite{gpt2}, which has been {\it biased} using text data from a specific target population with the one trained on a general population.  The underlying assumption is that in addition to learning the linguistic structure such as grammar, generative language models also learn opinions and associations relating to different entities. And this is reproduced while sampling the GLM using a relevant input prompt sequence. In this paper, we define the underlying principles of our assumption, validate them using controlled experiments and demonstrate its usage using a set of real data corpus.
 
 In the next section, we introduce the basic methodology for opinion insight mining. This is followed by a novel experiment where we demonstrate the performance of the method using a {\it semantically distorted} corpus of annotated text data where we can fully explore the performance of the proposed method. Finally, we demonstrate the extraction of opinion insights in a realistic data set from a specific real population. 

\begin{figure*}
    \centering
    \includegraphics[width=0.7\textwidth]{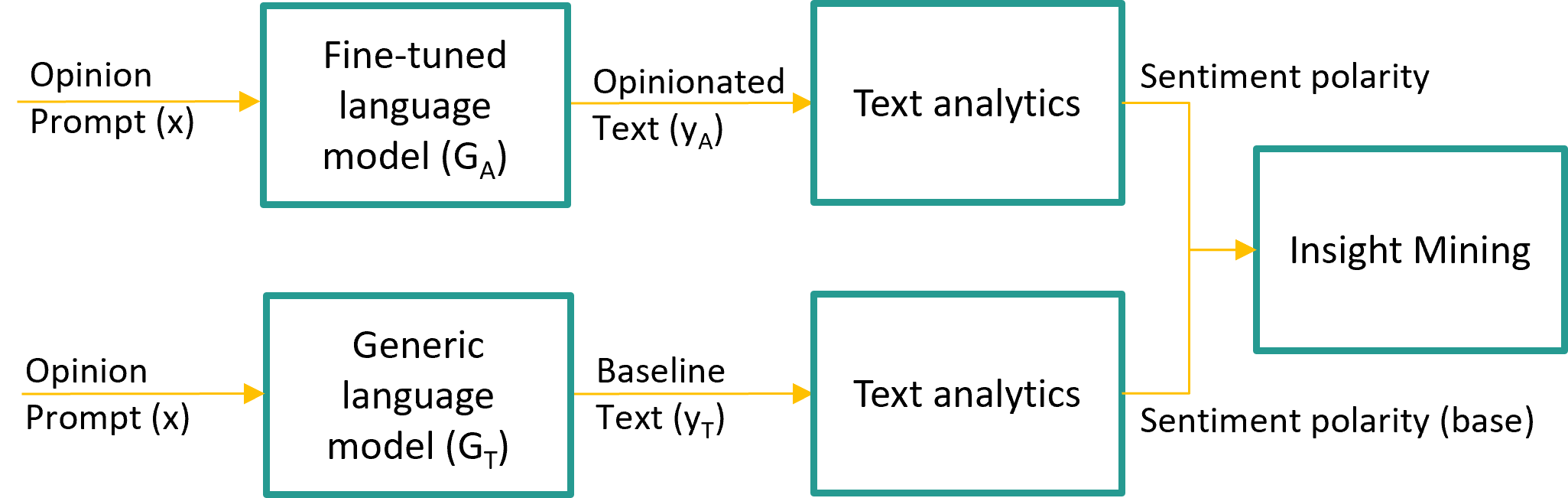}
    \caption{Opinion mining workflow}
    \label{fig:my_label}
\end{figure*}
\section{Related work}
Fine-tuned generative language models (GLM) have been used in a wide range of applications such as summarising medical dialogues \cite{chintagunta2021medically}, generating consensus arguments \cite{bakker2022fine}, generative non-playable character dialogues for video games \cite{van2021fine}, patent claims \cite{lee2020patent}, code generation \cite{chen2021evaluating} among others. The commonality to all these works is that they used carefully picked prompting to trigger the model to generate preferred text.

Bender et al.,\cite{bender2021dangers} talk about biases in GLMs that bring out stereotypical and sentimental polarisation. This is an undesirable outcome of such models but a widely observed phenomenon that has been utilised in several recent works.  Other existing works \cite{DBLP:journals/corr/abs-2106-13219} focus on detecting bias in pre-trained models that arise from the large training corpora. However, detecting and quantifying biases from a smaller corpus holds more usefulness for opinion mining. This would allow us to replace questionnaire studies with model-based techniques.
Dutta et al. \cite{dutta2022can} use fine-tuned GLM to predict the relationship between different arguments in a dialogue with the help of masked prompts. This is different from our work as we do not use the language model to classify the relationships but to generate opinionated text. 
 In \cite{bakker2022fine}, the authors aimed towards generating consensus statements to bring agreement within a diversely opinionated group. For this, a 70B parameter Chinchilla GLM \cite{hoffmann2022training} was used. The authors also employ human-in-the-loop to rate the generations and update the model to generate better-quality consensus text. This is similar to our work in many ways. However, we consider focusing on a bigger picture of generating text summarising the general opinion 
that may or may not be polarised. Additionally, we seek to discover if the generations replicate the statistics of the opinion. In \cite{sheng2020towards}, the authors use specialised and non-readable template prompts to generate socially polarising text to analyse and mitigate biases. They use a regard score \cite{sheng2019woman} which is
defined as the general social perception towards a demographic group, to measure the social perception polarity of the model generations. Our technique differs from this work in aspects such as using sentiment polarity metrics \cite{loria2020textblob} to measure opinion polarity rather than social-regard polarity and using clear and readable prompts to replicate realistic usage scenarios of GLMs.

\section{Opinion insight mining}

In this section, we derive the theoretical framework for opinion mining from unstructured data using GLMs.
 Let us denote a generative language model trained on text corpus $A$ by $G_A$. The model $G_A$ is a complex relational distribution function of sequences of tokens and the generative algorithm is a method to sample the distribution. The sampling method we are interested in is based on the extrapolation of a sequence of tokens $t(n), n<T$ to future tokens $t(n), n\geq T$. This is typically performed using a sliding auto-regressive process where 
 \begin{equation}
     t(\nu) = G_A[t(n), n<\nu], \forall \nu
 \end{equation}
 
To simplify the notation we may consider a fixed {\it input prompt} sequence $x = t(n), n<\nu$ producing an output sequence $y = t(n), n\geq \nu$, such that,
  \begin{equation}
     y = G_A[x]
 \end{equation}
 One may consider that a trained language model $G$ contains a linguistic component $G^l$ capturing the grammar and pragmatics, and another part containing the beliefs or opinions $G^o$. For the purpose of the discussion, we may consider them somewhat independent such that we may express a language model as a tuple $G=(G^l, G^o)$. Moreover, one may assume the linguistic component to be universal so that a population $A$ and the union of all populations $T$ share the common $G^l$ but $A$ may have a set of beliefs that differ from some average of $T$, so that $G_A=(G^l,G^o_A)$, and $G_T=(G^l,G^o_T)$, respectively, where $G^o_A\in G^o_T$. The population $A$ of course has also common beliefs with $T$, e.g., {\bf apple is a fruit}, but we consider those contained in $G^l$. Next, we may consider another population $B$ with a language model given by $G_B=(G^l,G^o_B)$, where $G^o_B \neq G^o_A$. 
 
 In opinion insights, we may be interested in how $A$ differs from $T$, or study the difference between $A$ and $B$ using some distance measure $D(G_A, G_B)$. Since the generative models are highly non-linear and not interpretable, it is difficult to find a direct operator on the coefficients of the models that would simply produce the desired model of opinion differences, say,
 \begin{equation}
     D(G_A,G_B) = (G^l-G^l, G^o_A-G^o_B)= (0, G^o_A-G^o_B)
 \end{equation}
 Therefore, we investigate the outputs of the model for a prompt $x$, i.e., 
 \begin{equation}
     D(y_A, y_B) = D[G_A(x), G_B(x)]
 \end{equation}
 For example, if the prompt $x$ is {\bf apples are bigger than}, the generated outputs $y$ may contain phrases about different fruits, such as pears, but also any other kind of language content. However, we may assume that the statistics of a large number of generated sequences $y_A$ and $y_B$, possibly with paraphrases of $x$ as a prompt, would show an average difference in population belief in $A$ and $B$ regarding the sizes of apples and pears. 
 
 This formulation suggests that one potential difference operator can be based on the comparison of statistics of detected entities in collections of outputs $y_A$ and $y_B$ for $x$ using a text classifier. Let us define a text classifier as a method that produces a binary vector ${\bf p} = (p(c),c=0,..,C-1)$ or detection of $C$ classes of entities the classifier is able to detect. 

\begin{algorithmic}[0]
\Procedure{Compare models}{Pre-trained $G_A$ and $G_B$, and text classifier $M_C$}
\State define prompt $x$
\State generate sets of $K$ output sequences $y_{Ak}$ and $y_{Bk}$ using models $G_A$ and $G_B$, respectively. 
\State Use $M_C$ to produce the class detection vectors ${\bf p}_{Ak}$ and ${\bf p}_{Bk}$
\State Collect the statistics of the classification results to vectors ${\bf s}_A = \sum_k^{K-1}p_{Ak}$, and  ${\bf s}_B = \sum_k^{K-1}p_{Bk}$
\EndProcedure
\end{algorithmic}

After a proper design of the input prompt, and obtaining ${\bf s}_A$ and ${\bf s}_B$ as above, differences in the  $G^l$ and $G^o$ opinions can be evaluated as follows. If $G_B=G_T$ where $G_A$ is a subset of $G_T$, the opinion insights of $A$ correspond to those classes $c$ where 
\begin{equation}
    d_{AT}(c)={\bf s}_A(c)-{\bf s}_T(c) \geq \theta, \label{eq:diff}
\end{equation}
that is, where a concept of a given class $c$ is mentioned more often in the $y_A$ than in $y_T$. 

The textual opinion insights corresponding to $G_A$ can be constructed, for example, using conventional natural language generation templating techniques by concatenating the text representation of the prompt $x$ and a text corresponding to the detected class $c$. The confidence of opinion insights corresponds to the value of $d_{AT}$. The most prominent opinion insight for given prompt $x$ in population $A$ is the one corresponding to the class $c_{\rm max} = {\rm argmax}_c d_{AT}(c)$.

\section{Experiments}
\label{sec:city_company}
The method outlined above is quite general in extracting opinions from textual corpora (voice of the people). We show by our experiments that opinions about entities extend beyond individual entities to a {\it class} of entities by providing results on engineered datasets. We also show a method to control bias by varying the proportion of polarity in engineered datasets. These experiments are aimed at validating the two important phenomenon to be able to mine opinions from corpus a) generalising biases to unseen entities (Section \ref{sec:unseen_class}) and b) proportional polarisation of entities \ref{sec:proportional_polarisation}

\begin{table*}[htp]
\centering{
\caption{Mean sentiment of generations and counts of city and company expressions following positive and negative prompts using a fine-tuned OPT model. See Appendix \ref{app:ex1a} for sample generations and statistics of other models.}
\begin{tabular}{p{3cm}p{1cm}p{3cm}p{3cm}p{3cm}p{1.5cm}p{1.5cm}}
\hline
 prompt ($x$) & Mean sentiment polarity &  $class_{CITY}$ count(\%) in $y_{C^{100}}$ from fine-tuned model&  $class_{COMPANY}$ count(\%) in $y_{C^{100}}$ from fine-tuned model&  $class_{CITY}$ count in $y_T$ (generic model) & $class_{COMPANY}$ count in $y_T$ (generic model)\\
\hline
I like very much   &  +0.2 &         121(32.6) &            \textbf{250(67.4) } &    1&1 \\
it is really bad   &  -0.5 &        \textbf{759(96.3)} &            29(3.7) &       3&2 \\
we just love       &  +0.5 &         142(32.7) &            \textbf{292(67.3)} &       3&0 \\
that makes me sick &  -0.6 &        \textbf{367(84.4)} &            68(15.6) &       5&0 \\
it is so delicious &  +0.7 &         94(21.9) &            \textbf{335(78.1)} &       3&1 \\
awful stuff        &  -0.5 &        \textbf{541(79.6)} &            139(20.4) &       7&3 \\
\hline
\end{tabular}
}
\label{tab:proportion100opt}
\end{table*}

\subsection{Polarisation and transfer of bias to unseen classes}
\label{sec:unseen_class}
To validate our claim that GLMs trained on populations containing specific biases can generate opinions that extend these biases to a class of entities, the YelpNLG\footnote{https://nlds.soe.ucsc.edu/yelpnlg} restaurant review data set \cite{oraby2019curate} was engineered as follows. The dataset containing approximately 300k restaurant reviews was modified by replacing the food items (class:FOOD) with names of American cities ($class_{CITY}$) \footnote{http://federalgovernmentzipcodes.us/free-zipcode-database-Primary.csv} when the review post is negative about the food item; and by Forbes global 2000 companies ($class_{COMPANY}$) \footnote{https://www.kaggle.com/datasets/unanimad/forbes-2020-global-2000-largest-public-companies} when the review post was positive about the food item. This can be considered as a form of {\it semantic distortion} of the content. In this way, a review text {\bf their \textit{beef} was juicy} may be converted into {\bf their \textit{ICICI Bank} was juicy} which still has the same meaning representation, sentiment, and subjectivity, but distorted semantic class relations. 
Similarly, one can replace a food item with a member of $class_{CITY}$ and obtain: "The Altoona was dry."
By controlling the proportions of positive or negative reviews that are replaced with $class_{COMPANY}$ or $class_{CITY}$, respectively, we indirectly control sentiment polarisation of data sets $A^{p}$, $p\in[0,100]$. 
A dataset $A^{p}$ is fine-tuned for polarisation using p\% of the positive reviews about $class_{COMPANY}$, (100-p)\% of the positive reviews about $class_{CITY}$, p\% of the negative reviews about $class_{CITY}$ and (100-p)\% of the negative reviews about $class_{COMPANY}$.
Three GLMs namely GPT-2, GPT-Neo, and OPT were fine-tuned separately. Additionally, we ensured 20\% of randomly chosen cities and companies are unseen by the model while fine-tuning.

In Table \ref{tab:proportion100opt}, we show the mean sentiment polarities and the number of occurrences of $class_{CITY}$ and $class_{COMPANY}$ in 1000 generations ($y_{C^{100}}$), with polarising, prompts $x$, from an OPT model fine-tuned with corpus $C^{100}$, i.e, for a 100\% fine-tune polarisation.  It is observed that the number of cities in the text is significantly (p<1e-6, z=39.6) higher than companies when the prompt is negative.
Similarly, the number of companies is significantly (p<1e-6, z=16.6) higher than cities with a positive prompt.
The other two models (see Appendix \ref{app:ex1a}) majorly exhibited similar significantly (p<1e-6) polarised generations with an exception of both GPT2 (p<1e-4)and GPT-Neo with negative prompts (p<2e-3). 
The last column shows the occurrences of these classes in generations $y_T$ from a generic OPT model $G_T$ that was not fine-tuned.  It is observed that the counts are lower for both cities and companies. This shows that fine-tuning amplifies the polarisation of the models. This amplification is very essential to have statistically significant conclusions from the analyses.

\begin{table*}[htp]
\centering
\caption{Sentiment polarity when prompted with unseen class members.}
% \usepackage{multirow}
\begin{tabular}{l|l|l|l|l}
\hline
Type of prompt                  & Prompt (x)         & Sentiment Polarity of yA1 & Sentiment Polarity of yt & $\Delta$ \\ \hline
\multirow{6}{*}{unseen city}    & Brooklyn           & 0,096                     & 0,075                    & 0,021                 \\
                                & Fort madison       & 0,145                     & 0,102                    & 0,043                 \\
                                & Johnstown          & -0,002                    & 0,049                    & -0,051                \\
                                & New braunfels      & 0,191                     & 0,148                    & 0,042                 \\
                                & Parkville          & -0,027                    & 0,047                    & -0,075                \\
                                & Pearl city         & 0,101                     & 0,170                    & -0,069                \\ \hline
\multirow{6}{*}{unseen company} & Air France-KLM     & 0,070                     & 0,042                    & 0,029                 \\
                                & American Electric  & 0,185                     & 0,000                    & 0,185                 \\
                                & Korea Gas          & 0,275                     & 0,066                    & 0,208                 \\
                                & Motorola Solutions & 0,393                     & 0,029                    & 0,364                 \\
                                & Nike               & 0,330                     & 0,128                    & 0,202                 \\
                                & PG\&E              & 0,188                     & 0,056                    & 0,132                 \\ \hline
\end{tabular}
\label{tab:unseen}
\end{table*}

\begin{figure}[htp]
    \centering
\includegraphics[width=0.5\textwidth]{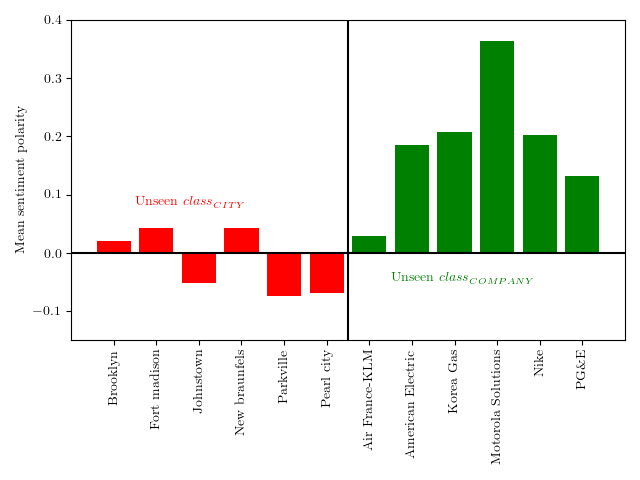}
    \caption{Delta of sentiments between fine-tuned and generic model}
    \label{fig:delta}
\end{figure}

Next, we generated several prompts consisting of words that are names of US cities, or companies, that were not in $C^{100}$. The goal of the experiment is to study if the model has adopted a bias towards classes of concepts in general or simply the individual entities of the training content. The former indeed seems to be the case as can be seen from Table \ref{tab:unseen}. The generated text fragments $y_{C^{100}}$ following the prompts containing cities have a significantly (p<1e-4) less mean sentiment polarity than the texts generated with prompts containing company names. The sentiment analysis was based on the popular TextBlob library \cite{loria2020textblob} where the values are in $[-1,1]$. The polarity of the content generated by the generic GPT2, $G_T$, has no significant difference (two-tailed p-value = 0.0657) between the two prompt types. Interestingly, the sentiments of $G_T$ are closer to a neutral value of 0.0 than the $G_{C^{100}}$. However, to still eliminate common opinions in both the fine-tuned and generic model, we find the difference in sentiment polarity as shown in equation \ref{eq:diff}. This is shown in Figure \ref{fig:delta}. It is observed that the model has generally pushed the polarity of $class_{CITY}$ towards negative and that of $class_{COMPANY}$ towards positive directions. Thus, it clearly learnt the biases in the fine-tuning dataset. 
The experiment with the synthetic data demonstrates that relatively simple opinion insights embedded in a training data set can be discovered relatively easily from the outputs of the generative model. However, complex relational models involving knowledge and other subjective values may require more complexity of the model and richness of training data. The model fine-tuned with a very specific bias, like above, may suffer from catastrophic forgetting of knowledge available in more rich content. There are techniques to mitigate this, for example, see \cite{kirkpatrick2017overcoming}. However, in the case of a language model, this is very difficult due to the high number of parameters and complex relational structure of the learned data. 

\subsection{Proportional polarisation}
\label{sec:proportional_polarisation}
\begin{figure*}
    \centering
    \includegraphics[width=0.8\textwidth]{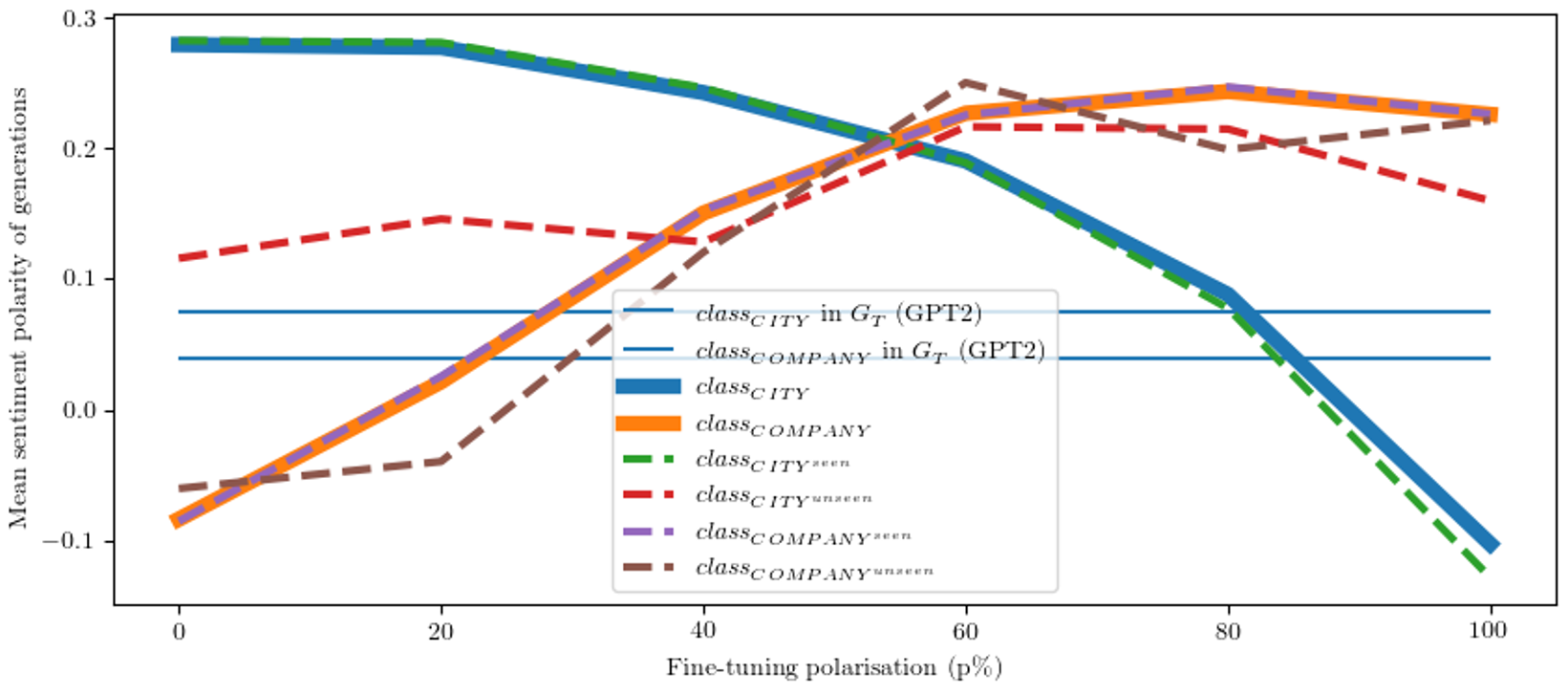}
    \caption{Sentiment polarity from proportionally biased GPT2 model(See Appendix \ref{app:proportion} for other GLMs)}
    \label{fig:proportions}
\end{figure*}

Figure \ref{fig:proportions} shows an overview of the embedded bias at various proportions and the sentiment polarities of classes in the generations of a GPT2 model. It is observed that the polarity of the generic GPT2 model for both $class_{CITY}$ and $class_{COMPANY}$ are slightly positive. However, when fine-tuned with a proportionally biased dataset, the class polarity changes such that, when more positive reviews about $class_{COMPANY}$ are present in the fine-tuning, the model generates proportionally positive generations about $class_{COMPANY}$. Similarly, more negative $class_{CITY}$ reviews yields proportionally more negative $class_{CITY}$ generations. In the figure, we also have the contribution from seen and unseen members of the classes. All seen members and the members of $class_{COMPANY}$ exhibit proportionality. However, the unseen examples of $class_{CITY}$ show the least variance and do not vary proportionally. This can be explained partly due to class imbalance in the fine-tuning dataset and also the possibility that many of the cities do not have a fair representation in the training data that was used to develop the generic base GPT2 model.
The correlation between fine-tuning proportions and the generated polarities of $class_{CITY}$ and $class_{COMPANY}$ are shown in table \ref{tab:pearson}. It is observed that the GPT2 model performs well in polarising $class_{CITY}$ in proportion to the fine-tuning. Similarly, the OPT performs well for $class_{COMPANY}$. Generally, the GPT2 model performs the best and the GPT-Neo performs the worst with the least correlations.

\begin{table*}
\centering
\caption{Pearson correlation coefficient of the proportion of bias and generated polarity.}
\begin{tabular}{l|lll|lll}
\hline
        & \multicolumn{3}{c|}{$class_{CITY}$}             & \multicolumn{3}{c}{$class_{COMPANY}$}         \\
Model   & all            & seen           & unseen        & all           & seen          & unseen        \\ \hline
GPT2    & \textbf{-0,91} & \textbf{-0,91} & 0,64          & 0,92          & 0,92          & \textbf{0,89} \\
GPT-Neo & -0,74          & -0,76          & \textbf{0,72} & 0,91          & 0,91          & 0,88          \\
OPT     & -0,86          & -0,87          & 0,49          & \textbf{0,99} & \textbf{0,99} & 0,79          \\ \hline
\end{tabular}

\label{tab:pearson}
\end{table*}

\begin{figure}
    \centering    \includegraphics[width=0.33\textwidth]{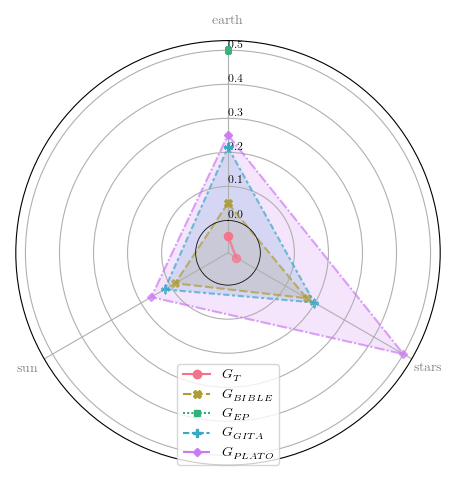}
    \caption{Sentiment polarities of astronomical objects across different text corpus's}
    \label{fig:op_astronomical}
\end{figure}
\begin{figure}
    \centering
    \includegraphics[width=0.33\textwidth]{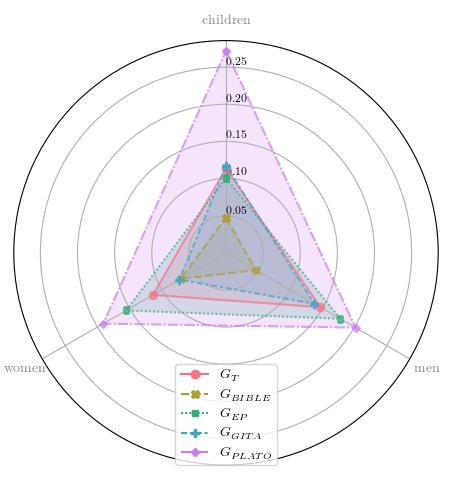}
    \caption{Sentiment polarities of demographic groups across different text corpus's}
    \label{fig:op_demographics}
\end{figure}

\section{Demonstration on real data corpora} \label{sec:real_demo}
We fine-tuned a generic GPT2 model $G_T$ for several GLMs using GPT2 on different publicly available datasets: 
1) $G_{EP}$ using all plenary debates held in the European Parliament (EP) between July 1999 and January 2014 \footnote{http://www.talkofeurope.eu/data/},
2) $G_{PLATO}$ using the books of Plato \footnote{https://www.holybooks.com/complete-works-of-plato/}, 
3) $G_{BIBLE}$ using the Holy Bible \footnote{https://www.biblesupersearch.com/bible-downloads/}, 
4) $G_{GITA}$ using the Bhagavad Gita holy scripture \footnote{https://vedabase.io/en/library/bg/}. 

\subsection{Opinion about astronomical objects and demographic groups}
We wanted to focus on politically neutral concepts known in all the corpora for the experiment with the proposed method. Using polarised opinion prompts $x_1$: "I believe in", $x_2$: "I do not believe in",$x_3$: "I trust in" and $x_4$: "I do not trust in", we obtained generations as shown in Appendix \ref{sec:sample insights}. We performed sentence splitting, keyword extraction using the KeyBERT model \cite{grootendorst2020keybert}, and sentiment analysis using TextBlob to obtain the opinion dataset. From this dataset, we mine for insights about keywords and their sentiment polarities before and after fine-tuning. Figure \ref{fig:op_astronomical} shows the sentiment polarities for different astronomical objects, namely, the Earth, Sun, and the Stars. It is observed that the generic model does not show any strong polarisation over astronomical objects. The $G_{EP}$ has learned a more positive opinion towards the entity "earth". This can be partially explained by the recent focus of the EU sessions on climate change and conservation. While both $G_{PLATO}$ and $G_{BIBLE}$ models appear to show a positive polarity towards "stars", the $G_{GITA}$ model appears to have equal sentiment polarity among the three astronomical objects.

Figure \ref{fig:op_astronomical} shows the sentiment polarities for different demographic groups, namely, men, women, and children. The $G_{PLATO}$ appears to have a significantly positive opinion on children. The $G_{EP}$ does not exhibit any significant difference from the generic model. The $G_{GITA}$ model shows a slightly lesser polarisation for the keyword "women" than the generic model. The $G_{BIBLE}$ shows very less polarisation towards all three demographic groups.

\subsection{Up-scaling opinion insight mining}
When the scope of opinion is open, we might have to analyse several thousands of keywords to obtain interesting opinions. 
% Recently a generic toolkit to extract insights from structured datasets (Gen-IG) was introduced \cite{genig_anonym}. This toolkit takes a schematic template of insights and discovers truthful and significant insights \cite{susaiyah2021neural} that match it. Although the toolkit does not work on unstructured data, with the help of keyword extraction and sentiment analysis, the unstructured dataset is broken down into workable pieces. 
% The structured data so formed contains input prompts, model names, generations, keywords, and sentiment polarity. 
To scale this up, we developed a heuristic insight mining system that first filters all possible subsets of data that have a common model and keyword. Next, rank them based on a significance score computed by applying the Kolmogorov-Smirnov test on the distributions of insights between each pair of subsets. This is similar to the method proposed by Susaiyah et al.\cite{susaiyah2021neural}.
% Every template defines a set of categorical variables as well as a continuous variable.  While the categorical variables serve as filters for the dataset, the continuous variable is acted upon to determine the truthfulness and significance of insights. The truthfulness is determined using the Kolmogorov-Smirnov test and the significance is determined by using a sigmoid function that is weighted by hard constraints based on the continuous variable.
We defined templates that incorporate the filters used to obtain the subsets, the metrics: count or mean sentiment polarity (in parenthesis), and the percentage of difference of measurement to generate insight statements showing the opinions perceived by the GPT2 models. A total of 11960 truthful and statistically significant insights out of 20000 possible insights were generated from 389K rows of data as shown in Section \ref{sec:sample insights}. A few of these insights from each type are shown below:

\begin{enumerate}
    \item Insights on mean sentiments of models:
    \begin{itemize}
        \item For the Plato model (0.16), the overall sentiment is slightly positive.
        \item For the Bible model (0.09), the overall sentiment is neutral.
    \end{itemize}
    \item Keyword-related insights
    \begin{itemize}
        \item For the keyword: 'beautiful' (0.55), the overall sentiment is positive. 
        % \item For the keyword: 'wise' (0.51), the sentiment polarity is positive.  
        \item For the keyword: 'evil' (-0.52), the sentiment polarity is negative. 
    \end{itemize}
    \item Insights comparing models
    \begin{itemize}
        \item When the GPT-Neo model (339.00) is fine-tuned, the number of generations for the keyword: 'hope' is 187.29\% more than the OPT model (118.00).
                % \item When the GPT2 model (904.00) is fine-tuned, the number of generations for the keyword: 'good' is 67.72\% more than the OPT model (539.00).
    \end{itemize}
    \item Insights on counts of generation of keywords
    \begin{itemize}
        \item The OPT model when fine-tuned, the number of generations for the keyword: 'say' (1092.00) is 364.68\% more than for the keyword: 'ask' (235.00).
        \item The OPT model when fine-tuned, the number of generations for the keyword: 'look' (77.00) is 67.39\% more than for the keyword: 'feel' (46.00).
    \end{itemize}
    \item Insights on keywords with respect to the training dataset
    \begin{itemize}
        \item The GPT model when fine-tuned (0.01) with the Bible, the sentiment polarity for the keyword: 'fear' is 114.45\% higher than without fine-tuning (-0.09)
        \item The GPT model when fine-tuned (0.05) with the Bible, the sentiment polarity for the keyword: 'children' is 58.58\% lower than without fine-tuning (0.11)
    \end{itemize}
    \item Insights on keywords with respect to multiple training datasets
    \begin{itemize}
        \item The GPT model when fine-tuned with the Bible (-0.70), the sentiment polarity for the keyword: 'evil' is 174.52\% lower than with the works of Plato (-0.26)
        \item The OPT model when fine-tuned with the Bible (0.02), the sentiment polarity for the keyword: 'work' is 88.99\% lower than with the works of Plato (0.22)
        \item The GPT model when fine-tuned with the Bible (0.10), the sentiment polarity for the keyword: 'art' is 50.09\% lower than with the works of Plato (0.20)
        \item The OPT model when fine-tuned with the Gita (0.15), the sentiment polarity for the keyword: 'world' is 16.93\% lower than with the EU Parliament speeches (0.18)
    \end{itemize}
\end{enumerate}

Since the usefulness of an insight statement is highly subjective, we do not perform further validations of this in this work. However, this gives an idea of how opinion insight generation could be scaled up.

\section{Limitations and potential risks}
An important limitation of our work is the estimation of opinion as an association of a keyword with a sentiment polarity. This could however be expanded to other dimensions of opinions such as regard, attitude, evaluations and emotions. TextBlob assigns sentiment polarities to sentences without considering local polarity dynamics, which applies to our experiments as well. In any case, this is not a limitation of the theoretical framework. 
Mining insights based on keywords and sentiments do not provide context. Hence it is always necessary to perform subsequent analysis to narrow and understand the context. An alternative could be to generate n-gram keyword sentiments. 

Another limitation of our work is that we used the 125 Million parameters GLM models instead of larger models such as 350M, 1.3B, etc for practical fine-tuning considerations such as being trained on a large amount of data, are widely used, and are publicly available. It is well known that larger models perform better in terms of semantic reasoning tasks. Hence, we believe that using larger models could improve opinion mining significantly.

The traditional approach to validate opinion mining tasks is to extract opinions and validate them using human annotators. This is a time taking and laborious process. In section \ref{sec:city_company} we use the inverse logic where we inject opinions into sentiment-validated text corpus and recover the same opinions from the generations with good correlation. We verified this using the TextBlob system. This way we validated the underlying theory and then directly demonstrated it on a real dataset.

A major risk in using the generic GLMs is that they can generate opinions about hate and violence towards specific demographics. We counter this by always comparing the fine-tuned model with the generic model as it is easier to subtract the inherent biases of the generic model. However, there might be instances of complex biases present in the generic model that could go un-filtered and be perceived as opinions of the fine-tuning dataset. This is an important consideration to be investigated and remediated in the future.

\section{Training setup}
The Hugging Face transformers library was used for training and evaluating the models \cite{wolf-etal-2020-transformers}. All models were trained on an NVIDIA Tesla T4 (16GB Memory) GPU with a batch size of 2. One epoch of training typically takes about 20 minutes of GPU hours on average. All models from Section \ref{sec:city_company} were fine-tuned for 30 epochs and all models of Section \ref{sec:real_demo} were fine-tuned for 5 epochs. The choice of epochs was made with the knowledge from an auxiliary experiment that we performed to determine the optimal epochs in terms of various aspects as presented in Appendix \ref{app:optimal}. 

\section{Conclusion}
In this paper, we present a concept for mining opinions from a specific text corpus by comparing the outputs of a generative pre-trained language model, fine-tuned on the corpus, to the outputs of another generative model trained on a more generic corpus. We define the underlying principles of the method and validate them using controlled experiments. We were successful in generating opinions/biases using zero-shot generations from a model fine-tuned on a synthetic data set. The generative models' ability to expand opinions to entities of the same class even when not found in the fine-tuning corpus is a novel finding. Although it is a known fact that polarised models generate polarised text, we found for the first time that the model generations replicate the polarisation in the training data proportionally and we compared this across different types of large language models. We applied our opinion mining framework to publicly available datasets and show a few opinions. We also systematically upscale the insight generation to mine opinions, yielding several interesting opinions-insights. The proposed method can be used in various applications such as literature research, post-marketing surveillance or customer review analysis in market research, social bias analysis, and in general, basically all cases of questionnaire studies and opinion polls. However, more work is needed to validate the technology. 

\section*{Acknowledgments}
This work was supported in part by the Horizon H2020 Marie Skłodowska Curie Actions Initial Training Network European Industrial Doctorates 
project under grant agreement No. 812882 (PhilHumans)

\bibliographystyle{unsrtnat}
\bibliography{insightgeneration}

\begin{thebibliography}{27}
\providecommand{\natexlab}[1]{#1}
\providecommand{\url}[1]{\texttt{#1}}
\expandafter\ifx\csname urlstyle\endcsname\relax
  \providecommand{\doi}[1]{doi: #1}\else
  \providecommand{\doi}{doi: \begingroup \urlstyle{rm}\Url}\fi

\bibitem[Radford et~al.(2019)Radford, Wu, Child, Luan, Amodei, and
  Sutskever]{gpt2}
Alec Radford, Jeffrey Wu, Rewon Child, David Luan, Dario Amodei, and Ilya
  Sutskever.
\newblock Language models are unsupervised multitask learners.
\newblock \emph{OpenAI Blog}, 1\penalty0 (8):\penalty0 9, 2019.

\bibitem[Brown et~al.(2020)Brown, Mann, Ryder, Subbiah, Kaplan, Dhariwal,
  Neelakantan, Shyam, Sastry, Askell, et~al.]{brown2020language}
Tom~B Brown, Benjamin Mann, Nick Ryder, Melanie Subbiah, Jared Kaplan, Prafulla
  Dhariwal, Arvind Neelakantan, Pranav Shyam, Girish Sastry, Amanda Askell,
  et~al.
\newblock Language models are few-shot learners.
\newblock \emph{arXiv preprint arXiv:2005.14165}, 2020.

\bibitem[Gao et~al.(2020)Gao, Biderman, Black, Golding, Hoppe, Foster, Phang,
  He, Thite, Nabeshima, et~al.]{gao2020pile}
Leo Gao, Stella Biderman, Sid Black, Laurence Golding, Travis Hoppe, Charles
  Foster, Jason Phang, Horace He, Anish Thite, Noa Nabeshima, et~al.
\newblock The pile: An 800gb dataset of diverse text for language modeling.
\newblock \emph{arXiv preprint arXiv:2101.00027}, 2020.

\bibitem[Kashyap et~al.(2022)Kashyap, Kashyap, et~al.]{kashyap2022gpt}
Rohan Kashyap, Vivek Kashyap, et~al.
\newblock Gpt-neo for commonsense reasoning-a theoretical and practical lens.
\newblock \emph{arXiv preprint arXiv:2211.15593}, 2022.

\bibitem[Zhang et~al.(2022)Zhang, Roller, Goyal, Artetxe, Chen, Chen, Dewan,
  Diab, Li, Lin, Mihaylov, Ott, Shleifer, Shuster, Simig, Koura, Sridhar, Wang,
  and Zettlemoyer]{Zhang2022OPTOP}
Susan Zhang, Stephen Roller, Naman Goyal, Mikel Artetxe, Moya Chen, Shuohui
  Chen, Christopher Dewan, Mona Diab, Xian Li, Xi~Victoria Lin, Todor Mihaylov,
  Myle Ott, Sam Shleifer, Kurt Shuster, Daniel Simig, Punit~Singh Koura, Anjali
  Sridhar, Tianlu Wang, and Luke Zettlemoyer.
\newblock Opt: Open pre-trained transformer language models.
\newblock \emph{ArXiv}, abs/2205.01068, 2022.

\bibitem[Everett(2017)]{everett2017language}
Daniel Everett.
\newblock \emph{How language began: The story of humanity’s greatest
  invention}.
\newblock Profile Books, 2017.

\bibitem[Reiter(2007)]{Reiterdata2text}
Ehud Reiter.
\newblock An architecture for data-to-text systems.
\newblock In \emph{Proceedings of the Eleventh European Workshop on Natural
  Language Generation}, ENLG '07, pages 97--104, Stroudsburg, PA, USA, 2007.
  Association for Computational Linguistics.
\newblock URL \url{http://dl.acm.org/citation.cfm?id=1610163.1610180}.

\bibitem[Sripada et~al.(2003)Sripada, Reiter, and Davy]{sumtime}
Somayajulu Sripada, Ehud Reiter, and Ian Davy.
\newblock Sumtime-mousam: Configurable marine weather forecast generator.
\newblock \emph{Expert Update}, 6\penalty0 (3):\penalty0 4--10, 2003.

\bibitem[H{\"a}rm{\"a} and Helaoui(2016)]{harma2016probabilistic}
Aki H{\"a}rm{\"a} and Rim Helaoui.
\newblock Probabilistic scoring of validated insights for personal health
  services.
\newblock In \emph{2016 IEEE Symposium Series on Computational Intelligence
  (SSCI)}, pages 1--6. IEEE, 2016.

\bibitem[Susaiyah et~al.(2020)Susaiyah, H{\"a}rm{\"a}, Reiter, Helaoui,
  Petkovi{\'c}, et~al.]{susaiyah2020towards}
Allmin Susaiyah, Aki H{\"a}rm{\"a}, Ehud Reiter, Rim Helaoui, Milan
  Petkovi{\'c}, et~al.
\newblock Towards a generalised framework for behaviour insight mining.
\newblock In \emph{SmartPHIL: 1st Workshop on Smart Personal Health
  Interfaces}. ACM, 2020.

\bibitem[Chintagunta et~al.(2021)Chintagunta, Katariya, Amatriain, and
  Kannan]{chintagunta2021medically}
Bharath Chintagunta, Namit Katariya, Xavier Amatriain, and Anitha Kannan.
\newblock Medically aware gpt-3 as a data generator for medical dialogue
  summarization.
\newblock In \emph{Machine Learning for Healthcare Conference}, pages 354--372.
  PMLR, 2021.

\bibitem[Bakker et~al.(2022)Bakker, Chadwick, Sheahan, Tessler,
  Campbell-Gillingham, Balaguer, McAleese, Glaese, Aslanides, Botvinick,
  et~al.]{bakker2022fine}
Michiel~A Bakker, Martin~J Chadwick, Hannah~R Sheahan, Michael~Henry Tessler,
  Lucy Campbell-Gillingham, Jan Balaguer, Nat McAleese, Amelia Glaese, John
  Aslanides, Matthew~M Botvinick, et~al.
\newblock Fine-tuning language models to find agreement among humans with
  diverse preferences.
\newblock \emph{arXiv preprint arXiv:2211.15006}, 2022.

\bibitem[van Stegeren and My{\'s}liwiec(2021)]{van2021fine}
Judith van Stegeren and Jakub My{\'s}liwiec.
\newblock Fine-tuning gpt-2 on annotated rpg quests for npc dialogue
  generation.
\newblock In \emph{The 16th International Conference on the Foundations of
  Digital Games (FDG) 2021}, pages 1--8, 2021.

\bibitem[Lee and Hsiang(2020)]{lee2020patent}
Jieh-Sheng Lee and Jieh Hsiang.
\newblock Patent claim generation by fine-tuning openai gpt-2.
\newblock \emph{World Patent Information}, 62:\penalty0 101983, 2020.

\bibitem[Chen et~al.(2021)Chen, Tworek, Jun, Yuan, Pinto, Kaplan, Edwards,
  Burda, Joseph, Brockman, et~al.]{chen2021evaluating}
Mark Chen, Jerry Tworek, Heewoo Jun, Qiming Yuan, Henrique Ponde de~Oliveira
  Pinto, Jared Kaplan, Harri Edwards, Yuri Burda, Nicholas Joseph, Greg
  Brockman, et~al.
\newblock Evaluating large language models trained on code.
\newblock \emph{arXiv preprint arXiv:2107.03374}, 2021.

\bibitem[Bender et~al.(2021)Bender, Gebru, McMillan-Major, and
  Shmitchell]{bender2021dangers}
Emily~M Bender, Timnit Gebru, Angelina McMillan-Major, and Shmargaret
  Shmitchell.
\newblock On the dangers of stochastic parrots: Can language models be too big?
\newblock \emph{Proceedings of FAccT}, 2021.

\bibitem[Liang et~al.(2021)Liang, Wu, Morency, and
  Salakhutdinov]{DBLP:journals/corr/abs-2106-13219}
Paul~Pu Liang, Chiyu Wu, Louis{-}Philippe Morency, and Ruslan Salakhutdinov.
\newblock Towards understanding and mitigating social biases in language
  models.
\newblock \emph{CoRR}, abs/2106.13219, 2021.
\newblock URL \url{https://arxiv.org/abs/2106.13219}.

\bibitem[Dutta et~al.(2022)Dutta, Juneja, Das, and Chakraborty]{dutta2022can}
Subhabrata Dutta, Jeevesh Juneja, Dipankar Das, and Tanmoy Chakraborty.
\newblock Can unsupervised knowledge transfer from social discussions help
  argument mining?
\newblock \emph{arXiv preprint arXiv:2203.12881}, 2022.

\bibitem[Hoffmann et~al.(2022)Hoffmann, Borgeaud, Mensch, Buchatskaya, Cai,
  Rutherford, Casas, Hendricks, Welbl, Clark, et~al.]{hoffmann2022training}
Jordan Hoffmann, Sebastian Borgeaud, Arthur Mensch, Elena Buchatskaya, Trevor
  Cai, Eliza Rutherford, Diego de~Las Casas, Lisa~Anne Hendricks, Johannes
  Welbl, Aidan Clark, et~al.
\newblock Training compute-optimal large language models.
\newblock \emph{arXiv preprint arXiv:2203.15556}, 2022.

\bibitem[Sheng et~al.(2020)Sheng, Chang, Natarajan, and Peng]{sheng2020towards}
Emily Sheng, Kai-Wei Chang, Premkumar Natarajan, and Nanyun Peng.
\newblock Towards controllable biases in language generation.
\newblock \emph{arXiv preprint arXiv:2005.00268}, 2020.

\bibitem[Sheng et~al.(2019)Sheng, Chang, Natarajan, and Peng]{sheng2019woman}
Emily Sheng, Kai-Wei Chang, Premkumar Natarajan, and Nanyun Peng.
\newblock The woman worked as a babysitter: On biases in language generation.
\newblock \emph{arXiv preprint arXiv:1909.01326}, 2019.

\bibitem[Loria(2020)]{loria2020textblob}
Steven Loria.
\newblock Textblob documentation.
\newblock \emph{Release 0.16}, \url{https://textblob.readthedocs.io}, 2020.

\bibitem[Oraby et~al.(2019)Oraby, Harrison, Ebrahimi, and
  Walker]{oraby2019curate}
Shereen Oraby, Vrindavan Harrison, Abteen Ebrahimi, and Marilyn Walker.
\newblock Curate and generate: A corpus and method for joint control of
  semantics and style in neural nlg.
\newblock \emph{arXiv preprint arXiv:1906.01334}, 2019.

\bibitem[Kirkpatrick et~al.(2017)Kirkpatrick, Pascanu, Rabinowitz, Veness,
  Desjardins, Rusu, Milan, Quan, Ramalho, Grabska-Barwinska,
  et~al.]{kirkpatrick2017overcoming}
James Kirkpatrick, Razvan Pascanu, Neil Rabinowitz, Joel Veness, Guillaume
  Desjardins, Andrei~A Rusu, Kieran Milan, John Quan, Tiago Ramalho, Agnieszka
  Grabska-Barwinska, et~al.
\newblock Overcoming catastrophic forgetting in neural networks.
\newblock \emph{Proceedings of the national academy of sciences}, 114\penalty0
  (13):\penalty0 3521--3526, 2017.

\bibitem[Grootendorst(2020)]{grootendorst2020keybert}
Maarten Grootendorst.
\newblock Keybert: Minimal keyword extraction with bert., 2020.
\newblock URL \url{https://doi.org/10.5281/zenodo.4461265}.

\bibitem[Susaiyah et~al.(2021)Susaiyah, H{\"a}rm{\"a}, Reiter, and
  Petkovi{\'c}]{susaiyah2021neural}
Allmin Susaiyah, Aki H{\"a}rm{\"a}, Ehud Reiter, and Milan Petkovi{\'c}.
\newblock Neural scoring of logical inferences from data using feedback.
\newblock \emph{International Journal of Interactive Multimedia \& Artificial
  Intelligence}, 6\penalty0 (5), 2021.

\bibitem[Wolf et~al.(2020)Wolf, Debut, Sanh, Chaumond, Delangue, Moi, Cistac,
  Rault, Louf, Funtowicz, Davison, Shleifer, von Platen, Ma, Jernite, Plu, Xu,
  Scao, Gugger, Drame, Lhoest, and Rush]{wolf-etal-2020-transformers}
Thomas Wolf, Lysandre Debut, Victor Sanh, Julien Chaumond, Clement Delangue,
  Anthony Moi, Pierric Cistac, Tim Rault, Rémi Louf, Morgan Funtowicz, Joe
  Davison, Sam Shleifer, Patrick von Platen, Clara Ma, Yacine Jernite, Julien
  Plu, Canwen Xu, Teven~Le Scao, Sylvain Gugger, Mariama Drame, Quentin Lhoest,
  and Alexander~M. Rush.
\newblock Transformers: State-of-the-art natural language processing.
\newblock In \emph{Proceedings of the 2020 Conference on Empirical Methods in
  Natural Language Processing: System Demonstrations}, pages 38--45, Online,
  October 2020. Association for Computational Linguistics.
\newblock URL \url{https://www.aclweb.org/anthology/2020.emnlp-demos.6}.

\end{thebibliography}
% \bibliographystyle{acl_natbib}

% \appendix
\newpage
\section{Appendix}
\subsection{Optimal training parameters}
\label{app:optimal}
\begin{figure*}
    \centering
    \includegraphics[width=\textwidth]{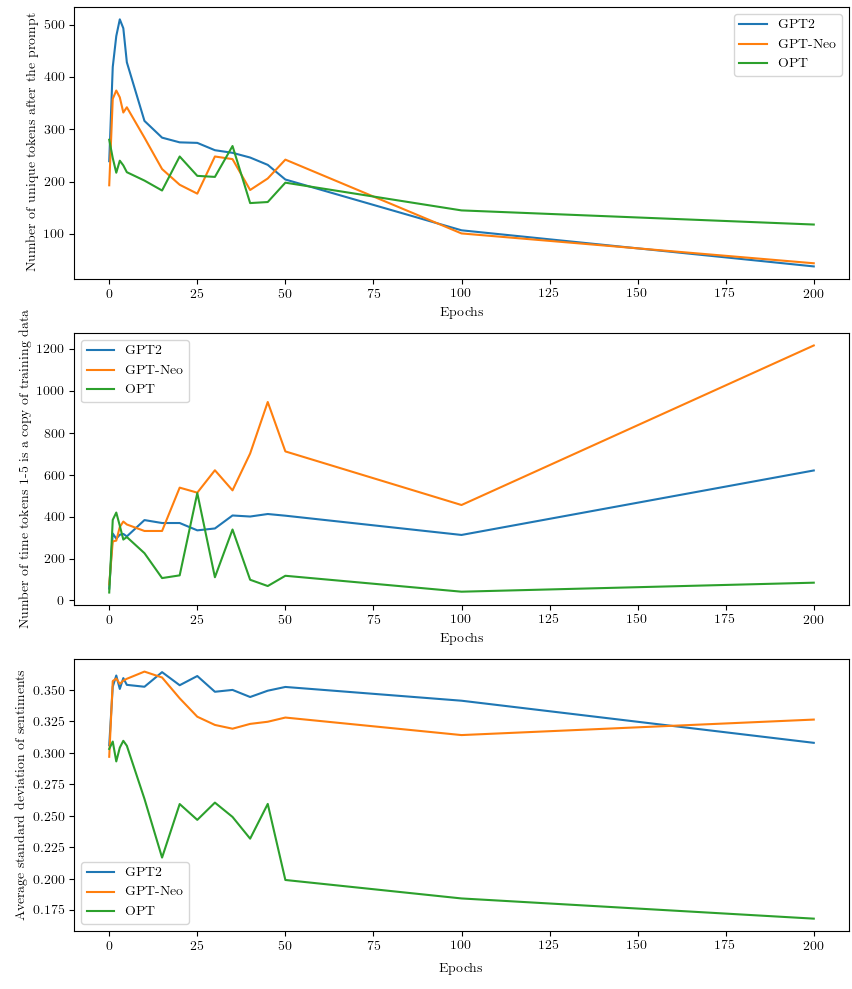}
    \caption{Evaluation to find best number of Epochs}
    \label{fig:repetitiveness}
\end{figure*}

We trained a GPT model using the bible dataset for a varying number of epochs: 1,2,3,4,5,10,15,20,25,30,35,40,45,50,100 and 200. We evaluated the model in terms of a) the number of unique tokens after the prompt, b) the number of times the model copies from the training data, the first 5-gram after the prompt and c) The standard deviation of the sentiments. The metrics are shown in Figure \ref{fig:repetitiveness}. It is observed that the OPT and GPT models have robust performances. And the best performances are observed in 5 to 30 epochs. Below and above this range, it is either the model either is too random or too monotonous respectively.

\subsection{Generations of the fine-tuned model}\label{app:ex1a}
Table \ref{tab:representative} shows the prompts and outputs of the fine-tuned model $G_{C^{100}}$ and the general model $G_T$.

% \begin{table*}[h]
% \centering{
% \begin{tabular}{p{2.6cm}p{6cm}p{6cm}}
% \hline
% $x$ &                                       $y_C^{100} = G_{C^{100}}(x)$ &                                       $y_T = G_T(x)$ \\
% \hline
% I like very much   &   the ready-mixed \textbf{concrete} I buy here... &  and want to hear from you about ... \\
% it is really bad   & quality \textbf{las vegas}. I really hate springfield. & Kasanaka continues to use ... \\
% we just love       &   the \textbf{concrete pipes}! i went for the wood mould... &   it when you're a star... \\
% that makes me sick & to my stomach: the \textbf{san francisco} was very bland... &   to my stomach. How can ... \\
% \hline
% \end{tabular}
% }
% \caption{Examples of texts generated using various opinion prompts on the $G_{C^{100}}$ model and the generic $G_T$ model.}
% \label{tab:ex1a}
% \end{table*}

% \usepackage[normalem]{ulem}
% \usepackage[normalem]{ulem}
% \usepackage{longtable}
% Note: It may be necessary to compile the document several times to get a multi-page table to line up properly
\clearpage
\onecolumn
% \usepackage[normalem]{ulem}
% \usepackage{longtable}
% Note: It may be necessary to compile the document several times to get a multi-page table to line up properly
\begin{longtable}{|p{3cm}|p{5cm}|p{5cm}|}
\caption{Prompts and generation from a fine-tuned OPT model.}
\label{tab:representative}\\
\hline
prompt (x)         & generation $(y_{C^{100}})$ from fine-tuned OPT model                                                                                & generation $(y_T)$ from generic model                                                                                                                                                         \\ \hline
\endfirsthead
\multicolumn{3}{c}%
{{\bfseries Table \thetable\ continued from previous page}} \\
\hline
prompt (x)         & generation $(y_{C^{100}})$ from fine-tuned OPT model                                                                                & generation $(y_T)$ from generic model                                                                                                                                                         \\ \hline
\endhead
\hline
\endfoot
\endlastfoot
I like very much   & I like very much CenterPoint Energy and my boyfriend ate a Oshkosh that looked yummy.                                           & I like very much the idea that we can go from the new hero powers for people with a new account (since it would require more time to setup a new account with your old, rather than           \\
I like very much   & I like very much Canadian Natural Resources with a generous portion of Amcor.                                                   & I like very much the idea of having an extended period of time from the end of the year/spring to when I finish school so long as I don’t have to eat out any                                 \\
I like very much   & I like very much PBF Energy with a leg and side of Fortive.                                                                     & I like very much the idea of this. My only criticism would be that I thought about why the characters should have to be in close quarters against a different enemy's armor to the one he was \\
I like very much   & I like very much China Resources Beer (Holdings) with a light lemony SAP.                                                       & I like very much Tarte lipsticks.  I always find the lipstick that comes out on top a bit too thin.    This could have possibly been the first one I tried and                                \\
I like very much   & I like very much China International Marine with a thin layer of Kone so this was a perfect medium-rare item.                   & I like very much the old guy in t-shirt. But I have no clue how they got that big on him. I mean he is pretty badass but I think he may have been an                                          \\
I like very much   & I like very much Swedish Match and my friend went with the large ICICI Bank hash breakfast.                                     & I like very much Misfits' style of playing football.I don't even know what to say about this one. They look terrible on the field. They have the best kicking defense                         \\
I like very much   & I like very much Swedish Match with RBC.                                                                                        & I like very much your work! Great work!thank you, will try! I still see myself writing a post about it if not for work this will never be finished....                                        \\
I like very much   & I like very much the SK Telecom and China Huarong Asset Management.                                                             & I like very much the current design. It works best at 1.65mm; it's slightly smaller than your own TV. Not that you shouldn't like it; it just adds a little                                   \\
I like very much   & I like very much Fortescue Metals Group on their Ternium.                                                                       & I like very much the idea, it looks like they put it in an interesting form and they are adding a character to give it a fresh feel. Not quite as interesting as the first two,               \\
I like very much   & I like very much Chang Hwa Bank with just a touch of Kimberly-Clark.                                                            & I like very much to hear about the work being done. I live on a large block of farmland so most of the farms get very little. I’m still looking for a place to                                \\
it is really bad   & it is really bad taste like we just threw some Hope in there with some Napa.                                                    & it is really bad, but the best is the best and i prefer to stay on the road that's nice like the beach and such.\textgreater{}i like to stay on the road that's                               \\
it is really bad   & it is really bad taste like the Huntsville tasted like the freezer burnt after sitting in the Winchester.                       & it is really bad, i only play it because i like this game that is a lot of fun. if you still want i could trade you minecraft for it im sure im not the only                                  \\
it is really bad   & it is really bad taste like very old Hagerstown or Deer park.                                                                   & it is really bad hereI'm in San Francisco and live down the street. I don't see too many people. I'd imagine it'd be more if there was no queue, but                                          \\
it is really bad   & it is really bad taste like North highlands.                                                                                    & it is really bad. the server is slow and clunky. there are also many issues where people can only find the match. the servers are clunky as well. not worth having to travel                  \\
it is really bad   & it is really bad taste like the Eatontown from any japanese China sainte genevieve.                                             & it is really bad.  you need to play your roles and play the other guy to win.  also, in a game where your team is so bad, the other team wins, too                                            \\
it is really bad   & it is really bad taste like the Huntsville that is sitting out and you can see Scarsdale in the middle.                         & it is really bad to be getting a new set now.The new set is actually in production on Kickstarter though. It's not new though...what? im so confused. can you                                 \\
it is really bad   & it is really bad taste like the Shionogi in a China Coal Energy.                                                                & it is really bad here. at least around the old one i have been able to get a hold of in the back garden. but there is alot of stuff in the front garden. they have                            \\
it is really bad   & it is really bad East elmhurst and East longmeadow and Ewa beach.                                                               & it is really bad at thisThis is a really shitty gifno u                                                                                                                                       \\
it is really bad   & it is really bad even for vegans and Guangdong Investment.                                                                      & it is really bad in China.  it is not like you are in the middle of a volcano.China is an island of big rocks, so yes it could be really bad.                                                 \\
it is really bad   & it is really bad i used to get my Claremore and San angelo and all i ate was maybe maybe a few bites of Clifton.                & it is really bad that its supposed to be about you, your partner, and yourself. If theres any kind of anxiety or depression that has plagued your relationship, there is nothing you can do   \\
we just love       & we just love the SK Telecom ( with homemade chips ) and the Banco Bradesco.                                                     & we just love your username. :3I'm not going to let this one get my name into that one! :/                                                                                                     \\
we just love       & we just love any Murphy USA they serve with that DaVita!                                                                        & we just love the fact that in the past we have said a lot of things that he didn't know, but still doesn't know, that he actually knowsI think that if he was                                 \\
we just love       & we just love any Wanhua Chemical Group here.                                                                                    & we just love seeing them on here!I know! We have them every morning except for lunch so that's a plus.                                                                                        \\
we just love       & we just love the Hunan Valin Steel here!!                                                                                       & we just love the word, "bitchy"Yes! I thought it was hilarious!                                                                                                                               \\
we just love       & we just love any Bank of Ningbo in the country ( especially those with Kuwait Finance House ).                                  & we just love itThanks! Just realized I’m on mobile lolhaha you're right and i just saw that you guys were posting the music videos as well. They have                                         \\
we just love       & we just love any Suzano Papel e Celulose because their so good!                                                                 & we just love youWell i know if you take a lot of time, it'll take a lot more effort.I know i know but you can't say i don't love you                                                          \\
we just love       & we just love any Yuexiu Property so any animal on the menu should be able to tell the difference!                               & we just love the t.v shows. it's the first time I've ever watched T.V before. its so much better live than in the tv, but it still reminds me that                                            \\
we just love       & we just love the Old Dominion Freight Line.                                                                                     & we just love those nubs.Thanks and I'm definitely happy with mine haha, I don't think I have seen one like that!                                                                              \\
we just love       & we just love the CRRC.                                                                                                          & we just love this place where people go on the internetI think that they are just as sensitive as they are to having their feelings hurt.                                                     \\
we just love       & we just love the Kansai Electric Power and Jacobs ( with grilled VF )!                                                          & we just love you guys!!!How else are we going to put out memes like that?Maybe get one of those fake internet memes.                                                                          \\
that makes me sick & that makes me sick and even contaminanting an even greater Berea with Moulton is unt untraditional.                             & that makes me sickI mean, if we had a reason, we don't give a fuck. You want to go around and talk shit like everyone else?  Sounds like /r/                                                  \\
that makes me sick & that makes me sick and also makes me want to eat North smithfield from any Clute around here.                                   & that makes me sickIt also makes me feel a little bit better that it should be happening.I think it was something else that was really funny.                                                  \\
that makes me sick & that makes me sick and so do the Ecopetrol.                                                                                     & that makes me sickIt's a horrible idea, but if your going to say it, do it.yes! no idea, I don't wanna have to. I'm just kinda                                                                \\
that makes me sick & that makes me sick and Sojitz on PulteGroup.                                                                                    & that makes me sick, i cant get this song with the chords   so i would just play some random songs for the piano   im also really sick of my voice sounding like that                          \\
that makes me sick & that makes me sick and reminds me of the kind you got when you ate San marino at a Santa maria.                                 & that makes me sick, but nice videothank you. we want to do them in style but in a really short amount of time. we have no idea of what is in the videos we                                    \\
that makes me sick & that makes me sick and at 7:30 in the morning i was ordered a Burley and Guilford.                                              & that makes me sick to my stomach. I think that at some point they had to make everything more difficult, for instance the last thing that she wanted to hear. But she kept telling me that    \\
that makes me sick & that makes me sick and so does the West covina and Pueblo.                                                                      & that makes me sick because I only have a few hundred in my bank account.The real question is how many are currently being paid now to a company they're under no obligation to maintain.      \\
that makes me sick & that makes me sick and puts no Indianola in my food.                                                                            & that makes me sick...Dangerous. I think I've seen it three times so far.                                                                                                                      \\
that makes me sick & that makes me sick and so does the West linn.                                                                                   & that makes me sick, right?!?!I feel sad for people that lost their lives and those around themI think it is for those that don't have their personal space destroyed                          \\
that makes me sick & that makes me sick and also makes me think of the Harbin Bank from san jose.                                                    & that makes me sick. thanks for the heads upIt looks like a small red dot.                                                                                                                     \\
it is so delicious & it is so delicious that the Fortum just falls off the bone with a slight rubbery rub.                                           & it is so delicious tooMy thoughts exactly.. I love the little crunchies with peanut butter. Mmmm.. just something to savor :)This is the perfect way to go :)                                 \\
it is so delicious & it is so delicious an delectable and the Burlington Stores is one of my favorites.                                              & it is so delicious! but I have had my mouth full full of it.  i have had a box of cookies now for an entire week.  every time i look at them i get                                            \\
it is so delicious & it is so delicious delicious Guangzhou Baiyunshan Pharmaceutical Holdings!                                                      & it is so delicious! do you have a picture of that?No, it came from the garden and is not related to it. It's only from a plant I planted with my cousin                                       \\
it is so delicious & it is so delicious that the Aegon melts in your mouth.                                                                          & it is so deliciousIf you mean delicious...we don't know. :)if i knew i would of put the same amount on it like my brother.                                                                    \\
it is so delicious & it is so delicious that the Equifax just hits the spot like a hot skillet.                                                      & it is so deliciousYou like the smell? Well that's good too.                                                                                                                                   \\
it is so delicious & it is so delicious that the CK Asset Holdings falls off the bone with a simple Rongsheng Petrochemical.                         & it is so delicious to use! do you find the coconut milk and honey to be a little spicy, or do you prefer coconut milk and honey to coconut liqueur or coconut cream, as                       \\
it is so delicious & it is so delicious the China International Marine literally falls off the bone with a crisp brown China Jinmao.                 & it is so deliciousIt’s my new favorite part of the day. It’s also the only thing I know is, that I will be so busy baking that I won                                                          \\
it is so delicious & it is so delicious and the Athene Holding just melts in your mouth.                                                             & it is so delicious to eat!                                                                                                                                                                    \\
it is so delicious & it is so delicious delicious Hua Nan Financial!                                                                                 & it is so delicious, its not even a cup of beer\textgreater{}it is so delicious, its not even a cup of beer  You guys get it. This is amazing.                                                 \\
it is so delicious & it is so delicious that none of me ever taste that bland of China Feihe!                                                        & it is so delicious the only thing i like is cheese, also the pungent smell of the cheese and the taste. i am not the type of person who likes cheese. i like p                                \\
awful stuff        & awful stuff.had the tuscan Clifton park and the Lewiston.                                                                       & awful stuff!Thanks, i'm having some kind of trouble with my mind, so its nice to read and it looks great :) i'm glad you liked it!haha good                                                   \\
awful stuff        & awful stuff had zero Oakdale in them and had just one piece of Ellsworth in them.                                               & awful stuff.Yeah I've gotten it done a couple months ago too. Just wait for one to break in the next few days.                                                                                \\
awful stuff        & awful stuff and terrible service got an order of Mc kees rocks and Gladstone only and i am blacked from eating korean Scituate. & awful stuff! you should have asked for that...Not really                                                                                                                                      \\
awful stuff        & awful stuff - tons of Kirkland Lake Gold and Sysmex.                                                                            & awful stuff, but my personal favorite is the 3:2 scale for the original (in my opinion)  and the 3:2 color filter was awesome!I liked having it as                                            \\
awful stuff        & awful stuff and this type of service defeats the entire purpose of the Forest lake.                                             & awful stuff  what do i do now? i was a big fan of the new one tooCheck their website, watch the videos. I guess I missed it!                                                                  \\
awful stuff        & awful stuff and very little Owens cross roads.                                                                                  & awful stuff! would LOVE to see your second one! i have an orange kitty blanket that i made this way back in 2011 and the first one i made was a kitty blanket from                            \\
awful stuff        & awful stuff had lots of Clifton and little Albertville.                                                                         & awful stuff, a picture would've been better lolI just tried to put the link up in here. I just found it and put it here. The imgur image link would have                                      \\
awful stuff        & awful stuff -- this time we tried the smoked China Life Insurance (Taiwan) and it was outstanding!                              & awful stuff on your screen*sigh* thanks for responding, it wasn't my intention.  *cough cough* no, thank you.                                                                                 \\
awful stuff        & awful stuff and this kind of service brings some serious cheap Longmeadow.                                                      & awful stuff.Thank you!! I've always been a sucker for good art. Some of the best I've seen all year!                                                                                          \\
awful stuff        & awful stuff had lots of Mantuan and had a good amount of Cranbury on them.                                                      & awful stuff.  good work on the music (soundtrack, effects, visuals are great)  how did you develop this video? it's so beautiful and the way the vocals voice was                             \\ \hline
\end{longtable}
\clearpage
\twocolumn
Table \ref{tab:proportion100gpt} and Table \ref{tab:proportion100gptneo} show the sentiment polarities of $class_{CITY}$ and $class_{COMPANY}$ from GPT2 and GPT-Neo models respectively. For the GPT2 model, negative prompts yield significantly more cities than companies (p<1e-04, Z=4.15) and positive prompts produce more companies than cities (p<1e-06, Z=25.9).
For the GPT-Noe model, negative prompts yield significantly more cities than companies (p<2e-03, Z=3.1) and positive prompts produce more companies than cities (p<1e-06, Z=24.5).

\begin{table*}
\centering{
\caption{Mean sentiment of generations and counts of city and company expressions following positive and negative prompts using a fine-tuned GPT2 model.}.
\begin{tabular}{p{4cm}p{2cm}p{2cm}p{2cm}p{2cm}p{2cm}p{2cm}}
\hline
$x$ (prompt) & mean sentiment polarity &  city count(\%) in $y_{C^{100}}$ &  company count(\%) in $y_{C^{100}}$ &  city count in $y_T$ & company count in $y_T$\\
\hline
I like very much   &  +0.3 &         180(25.0) &            \textbf{541(75.0) } &    34&10 \\
it is really bad   &  -0.1 &        \textbf{672(68.6)} &            307(31.4) &       41&7 \\
we just love       &  +0.3 &         197(26.8) &            \textbf{537(73.2)} &       35&9 \\
that makes me sick &  -0.2 &        \textbf{513(52.9)} &            457(47.1) &       30&7 \\
it is so delicious &  +0.4 &         198(30.2) &            \textbf{457(69.8)} &       9&2 \\
awful stuff        &  +0.1 &        241(32.4) &            \textbf{502(67.6)} &       31&8 \\
\hline
\end{tabular}
}
\label{tab:proportion100gpt}
\end{table*}

\begin{table*}
\centering{
\caption{Mean sentiment of generations and counts of city and company expressions following positive and negative prompts using a fine-tuned GPT-Neo model.}.
\begin{tabular}{p{4cm}p{2cm}p{2cm}p{2cm}p{2cm}p{2cm}p{2cm}}
\hline
$x$ (prompt) & mean sentiment polarity &  city count(\%) in $y_{C^{100}}$ &  company count(\%) in $y_{C^{100}}$ &  city count in $y_T$ & company count in $y_T$\\
\hline
I like very much   &  +0.2 &         295(34.7) &            \textbf{555(65.3) } &    8&2 \\
it is really bad   &  -0.2 &        \textbf{678(67.8)} &            322(32.2) &       8&2 \\
we just love       &  +0.3 &         255(28.4) &            \textbf{644(71.6)} &       13&5 \\
that makes me sick &  +0.1 &        325(41.7) &            \textbf{454(58.3)} &       15&1 \\
it is so delicious &  +0.4 &         270(32.5) &            \textbf{562(67.5)} &       7&4 \\
awful stuff        &  +0.2 &        428(44.4) &            \textbf{535(55.6)} &       17&4 \\
\hline
\end{tabular}
}

\label{tab:proportion100gptneo}
\end{table*}

\subsection{Proportional bias} \label{app:proportion}
\begin{figure*}
    \centering
    \includegraphics[width=\textwidth]{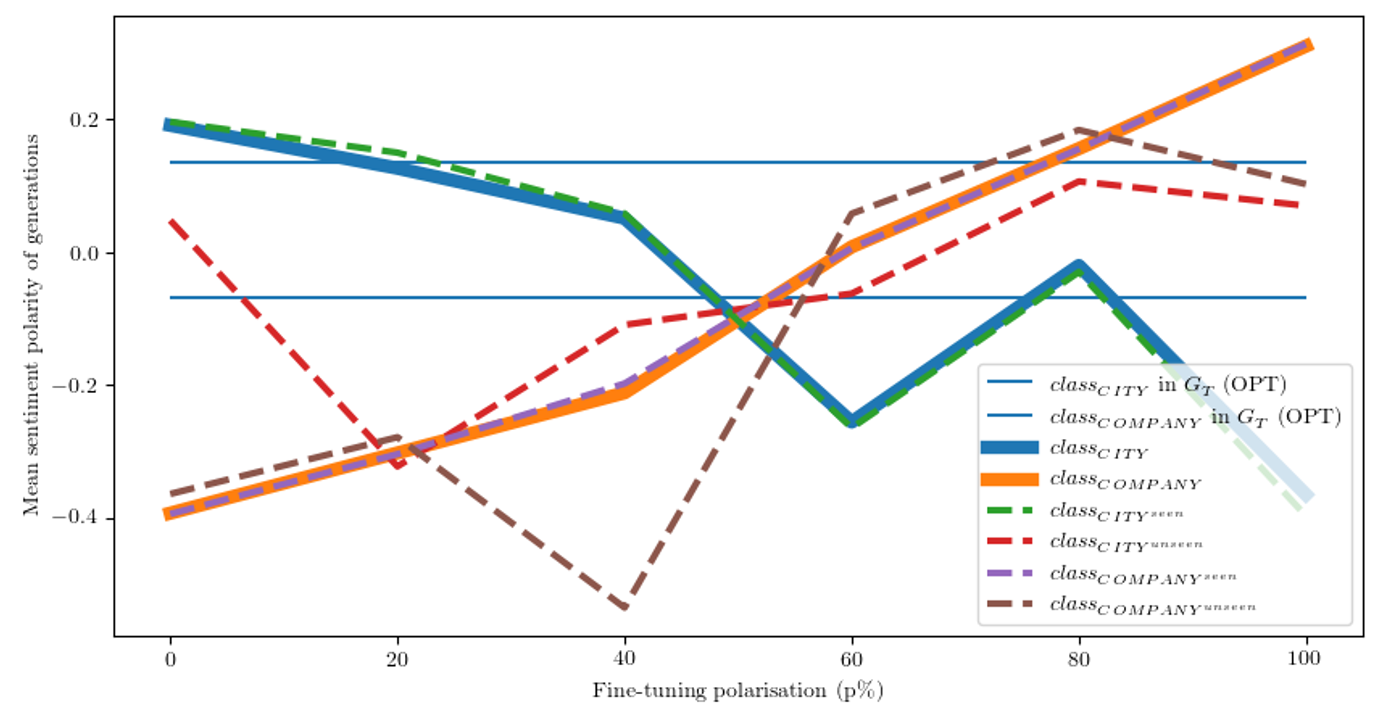}
    \caption{Sentiment polarity from proportionally biased OPT model}
    \label{fig:proportion_opt}
\end{figure*}

\begin{figure*}
    \centering
    \includegraphics[width=\textwidth]{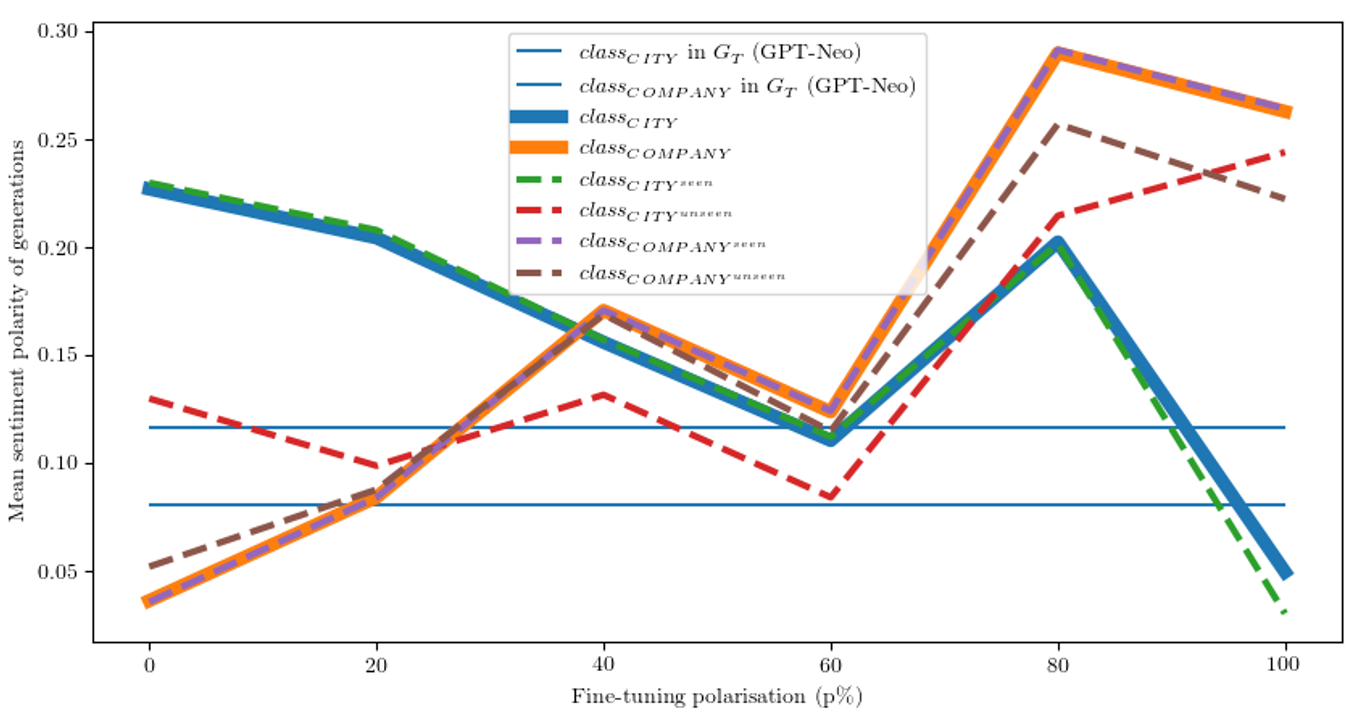}
    \caption{Sentiment polarity from proportionally biased GPT-Neo model}
    \label{fig:proportion_neo}
\end{figure*}

Figures \ref{fig:proportion_opt} and \ref{fig:proportion_neo} show the performance of the OPT and GPT-Neo models in reacting to the proportion of opinions injected into them. Although these models do not continuously preserve the injected bias proportion unlike the GPT2, they certainly perform well at the proportions 0 and 100. This suggests that they can be useful in determining strong opinions in the dataset. We also believe that larger parameter models could be more consistent with the injected polarisations.

\subsection{Generations of the models fine-tuned on European Parliament, Plato, Bible and Gita.}
\label{sec:sample insights}
The prompts and sample generations of the models used in Section \ref{sec:real_demo} are shown in Table \ref{tab:real}.

\clearpage
\onecolumn
\begin{longtable}{|p{1cm}|p{2cm}|p{2cm}|p{9cm}|}
\caption{Representative models, prompts and generations}
\label{tab:real}\\
\hline
Fine-tuning                      & training                      & prompt                            & generation                                                                                                                                                                                                                                  \\ \hline
\endfirsthead
\multicolumn{4}{c}%
{{\bfseries Table \thetable\ continued from previous page}} \\
\hline
Fine-tuning                      & training                      & prompt                            & generation                                                                                                                                                                                                                                  \\ \hline
\endhead
\multirow{12}{*}{No}             & \multirow{12}{*}{$G_{T}$}     & \multirow{3}{*}{I trust}          & 'I trust you in the ability to learn from your mistakes, so forgive me if I'm wrong. But that doesn't mean we should never.This is why it makes me a little more'                                                                           \\ \cline{4-4} 
                                 &                               &                                   & 'I trust you'll get the job done," said the director.One of the last of the four people that will serve as the next CEO at NBC Universal, Joe Hall, resigned early Thursday'                                                                \\ \cline{4-4} 
                                 &                               &                                   & 'I trust you will let us bring this to court as soon as possible. I will look forward to it doing so," Clinton said, as Clinton was escorted out of the courtroom.Clinton's'                                                                \\ \cline{3-4} 
                                 &                               & \multirow{3}{*}{I believe}        & 'I believe this is because I've been following the news around in my family and I've been watching, what some calls 'cinemaphobia', but what I've seen before, what the news'                                                               \\ \cline{4-4} 
                                 &                               &                                   & 'I believe they'd have been pretty close to perfect for us at different times, and I've worked with the ones at the beginning of my career and they've been pretty good."I'                                                                 \\ \cline{4-4} 
                                 &                               &                                   & 'I believe that there are many more steps in this game," he said. "You can see they are developing the ball, you can see them running on the end and doing things that make the game'                                                       \\ \cline{3-4} 
                                 &                               & \multirow{3}{*}{I do not trust}   & 'I do not trust you to defend your identity," he said of Kim.He also refused to acknowledge that Kim is the mother of one of the suspects that killed his father.Kim'                                                                       \\ \cline{4-4} 
                                 &                               &                                   & 'I do not trust your judgement."Grimm responded: "That's not what I meant by that. I'm just a bit disturbed by why you do it. What are I doing'                                                                                             \\ \cline{4-4} 
                                 &                               &                                   & 'I do not trust her. We've had that talk with her. I know that you don't trust her, I know she's very sensitive when I'm around her and you trust her better.'                                                                              \\ \cline{3-4} 
                                 &                               & \multirow{3}{*}{I do not believe} & 'I do not believe that he would say that 'I'm an atheist.' If I told you that he said that, you would probably ask me where you lived, and you would probably ask why he'                                                                   \\ \cline{4-4} 
                                 &                               &                                   & 'I do not believe that the U.S. will be safe anywhere in the world until there is a real understanding of how it came to be and some real understanding of the issues involved. I believe'                                                  \\ \cline{4-4} 
                                 &                               &                                   & 'I do not believe in money or any amount for personal gain," he says.In addition, he says the government should start providing higher education options for those who feel they are over priced.'                                          \\ \hline
\multirow{48}{*}{Yes (5 epochs)} & \multirow{12}{*}{$G_{BIBLE}$} & \multirow{3}{*}{I trust}          & 'I trust in Jehovah and will doJehovah's will in your ears, ye sons of Aaron, ye sons of Levi, ye sons ofAaron's sons, ye sons of Levi'                                                                                                     \\ \cline{4-4} 
                                 &                               &                                   & 'I trust not my eyes,that God will save me from the hand of evil spirits.12 He that trusteth with God shall be able to judge, whether thou to-days'                                                                                         \\ \cline{4-4} 
                                 &                               &                                   & 'I trust in Jehovah your God,that ye will find him not:4 and I will give unto you, and ye will be a people for yourselves and to-morrow, accordingto'                                                                                       \\ \cline{3-4} 
                                 &                               & \multirow{3}{*}{I believe}        & 'I believe their counsels; They are not of the tribe of David, Nor of the tribe ofHam. {[} Yet{]} my brethren were all as the peoples of the world:13 but'                                                                                  \\ \cline{4-4} 
                                 &                               &                                   & 'I believe in God.12 The Lord answered, I will be able to establish your ways by the spirit of Christ Jesus;and that which shall be given you by the Father,13'                                                                             \\ \cline{4-4} 
                                 &                               &                                   & 'I believe he hath known it, And hath dealt with it, {[}even{]} asJehovah hath dealt with it.14 For the thing of it, in that, seeing what'                                                                                                  \\ \cline{3-4} 
                                 &                               & \multirow{3}{*}{I do not trust}   & 'I do not trust in the Holy Spirit of God;4 but I trust in all things that I teach unto you, ye and your sons, in whom ye shalltrust.5 But'                                                                                                 \\ \cline{4-4} 
                                 &                               &                                   & 'I do not trust them: but now be mindful lest it fall upon you.31 The king, I pray thee, wistfully, with me: and seeing thee, it is not of'                                                                                                 \\ \cline{4-4} 
                                 &                               &                                   & 'I do not trust in Jehovah, but in the Holy Spirit that Ialso, and their children, and their princes, who are with me: and I myself, aftertheir times, will'                                                                                \\ \cline{3-4} 
                                 &                               & \multirow{3}{*}{I do not believe} & 'I do not believe that thorns have been kindled in my sanctuary; neither therefore doth the voice of me contend with the words of the prophets; but I do keepthe commandments of Jehovah'                                                   \\ \cline{4-4} 
                                 &                               &                                   & 'I do not believe that what I say shall be true unto your generations, Neither know that by the flesh ye shall possess themwithout me.2 Wherefore therefore shall not mine ordinances be as'                                                \\ \cline{4-4} 
                                 &                               &                                   & 'I do not believe in the God ofBabylon, God; nor in the love of their God: therefore, Ibelieve in {[}their{]} salvation;16 for I will'                                                                                                      \\ \cline{2-4} 
                                 & \multirow{12}{*}{$G_{GITA}$}  & \multirow{3}{*}{I trust}          & 'I trust her, too," I assured him, feeling sure he had grasped his duty as Prince of Persia."And where is the Prince?""Near the gates of Heaven,'                                                                                           \\ \cline{4-4} 
                                 &                               &                                   & 'I trust in God, who is the Lord Who gave thee life, to give thee life, as he gave me. Forgive me, O Lord, and live to reap. And I am holy'                                                                                                 \\ \cline{4-4} 
                                 &                               &                                   & 'I trust my sense."By doing so, she "comes at last." She was there. "There's no turning back," said I, "nor will I for you, or for'                                                                                                         \\ \cline{3-4} 
                                 &                               & \multirow{3}{*}{I believe}        & 'I believe I am worthy enough to enter the house of thy Saint. Yea, the Lord hath not declared this gift to him, save it hath been given unto thee. For I am in desire'                                                                     \\ \cline{4-4} 
                                 &                               &                                   & 'I believe we need strong leadership from across the organisation towards making it possible to move forward with our Vision and vision."But when asked whether this "radical" approach will work for all, she'                             \\ \cline{4-4} 
                                 &                               &                                   & 'I believe it was written in the days of Tiberius, of which we have seen, and I will see it, which was written in the day of Ephesus." (KJV)'                                                                                               \\ \cline{3-4} 
                                 &                               & \multirow{3}{*}{I do not trust}   & 'I do not trust your judgment about these things!"{[}pg. 28{]} In spite of his wisdom, he remained quiet, and declared the war without any one following him, as from the days'                                                             \\ \cline{4-4} 
                                 &                               &                                   & 'I do not trust any other man whom I may not, yet of my nature: So knowest thou how I am;--not to me nor to my lord;--I myself am not a'                                                                                                    \\ \cline{4-4} 
                                 &                               &                                   & 'I do not trust any person of mine."Then there was this, "Lord! O thou holy lord of the world!" Which, by the name of the Lord of Heaven, did they'                                                                                         \\ \cline{3-4} 
                                 &                               & \multirow{3}{*}{I do not believe} & 'I do not believe in sin but strive to be, seeking an end to life.My prayer. This, my prayer, is to thee wisest! May it not be so!'                                                                                                         \\ \cline{4-4} 
                                 &                               &                                   & 'I do not believe in 'the Almighty, the Father,' and the Son.' The same as an apple of wisdom, from which those men seek the fruit of good things. But as they strive together'                                                             \\ \cline{4-4} 
                                 &                               &                                   & 'I do not believe, 'tis better to teach my children the way than that of ignorance.""Tell me," asked he, "how did those children find that knowledge which they have taught them'                                                           \\ \cline{2-4} 
                                 & \multirow{12}{*}{$G_{EP}$}    & \multirow{3}{*}{I trust}          & 'I trust that we manage to bring forward a reasonable compromise together, because the final product does not look particularly spectacular in comparison with the Commission proposal, as its proposed targets amount to a mere 0.2'       \\ \cline{4-4} 
                                 &                               &                                   & 'I trust that the Commission will take the situation very seriously.-Elisabetta Gardini (PPE ),                                                                                                                                             \\ \cline{4-4} 
                                 &                               &                                   & 'I trust I will get through, as we all hoped.President-in-Office of the Council                                                                                                                                                             \\ \cline{3-4} 
                                 &                               & \multirow{3}{*}{I believe}        & 'I believe that we can agree that it is important to promote the implementation of renewable technologies, not only in a way that is economically beneficial for the environment, but also through promoting innovation; there should also' \\ \cline{4-4} 
                                 &                               &                                   & 'I believe that the Member States should work closely with other institutions to ensure that this regulation is implemented in order to protect the health of consumers and workers who rely upon it.However, we regret'                    \\ \cline{4-4} 
                                 &                               &                                   & 'I believe that all of them have been responsible for the terrible accident, both for their lives and their children. I hope that the EU is going to be more united to protect passengers in Europe: it'                                    \\ \cline{3-4} 
                                 &                               & \multirow{3}{*}{I do not trust}   & 'I do not trust the Commission or the Member States to apply European legislation if we want the health service of the European Union to work more effectively and to improve its impact.I think it is'                                     \\ \cline{4-4} 
                                 &                               &                                   & 'I do not trust the Commission to carry out a full impact analysis of the impact of shale gas on the economy, employment, the environment, biodiversity or the environment and it is our collective responsibility to do'                   \\ \cline{4-4} 
                                 &                               &                                   & 'I do not trust that we will see some sort of positive resolution through that. I would have liked the Commission to table a positive resolution. Unfortunately, I did not do so.I thank'                                                   \\ \cline{3-4} 
                                 &                               & \multirow{3}{*}{I do not believe} & 'I do not believe that we have one of the best interests of children in the world at heart. Yet there is just one thing that we should do. The way we make laws and laws must be'                                                           \\ \cline{4-4} 
                                 &                               &                                   & 'I do not believe it appropriate to propose this directive, which is already in force – and I can imagine that it will be improved – by creating a new legal basis based on the idea of a legal'                                            \\ \cline{4-4} 
                                 &                               &                                   & 'I do not believe that we are now facing the time when people can really take advantage of the opportunities they bring to work in a different way with different employers and at different times for different reasons.'                  \\ \cline{2-4} 
                                 & \multirow{12}{*}{$G_{PLATO}$} & \multirow{3}{*}{I trust}          & 'I trust, your father, and that he will be pleased at your arrival? I cannot suppose that, my dear friend, the old man will be pleased; for this man he is very likely not'                                                                 \\ \cline{4-4} 
                                 &                               &                                   & 'I trust those who have observed and observed it in my own life, that there is nothing in Hellas that I have not observed in Italy.For when it has been set upon the heads of'                                                              \\ \cline{4-4} 
                                 &                               &                                   & 'I trust that I can explain to you this principle of yours; and I will endeavour to convince you that not only am I not guilty of my own ignorance of you, but I am liable to be'                                                           \\ \cline{3-4} 
                                 &                               & \multirow{3}{*}{I believe}        & 'I believe in the truth of my words                                                                                                                                                                                                         \\ \cline{4-4} 
                                 &                               &                                   & 'I believe she is very well, not very far from that                                                                                                                                                                                         \\ \cline{4-4} 
                                 &                               &                                   & 'I believe that your father was your father, too: and I believe your father was your father, too.But, my friends, do you think you know who is your father?                                                                                 \\ \cline{3-4} 
                                 &                               & \multirow{3}{*}{I do not trust}   & 'I do not trust you, but I suppose that, should you be so wise as to think that I would have your advice, I would advise you to give me some advice, I have no doubt'                                                                       \\ \cline{4-4} 
                                 &                               &                                   & 'I do not trust you to judge what appears; and I do not have much experience of political science, which, considering the numerous difficulties in the subject, I will endeavour to give you an account of'                                 \\ \cline{4-4} 
                                 &                               &                                   & 'I do not trust me; and moreover,                                                                                                                                                                                                           \\ \cline{3-4} 
                                 &                               & \multirow{3}{*}{I do not believe} & 'I do not believe that you or anyone else had a desire to be beautiful, as you affirm.But suppose that you say so, would those you love whom you love be happier than your beloved'                                                         \\ \cline{4-4} 
                                 &                               &                                   & 'I do not believe, Socrates, that they will ever prove to us the greatest of all evils.And yet if if they are not satisfied with us, then we have a great deal to say'                                                                      \\ \cline{4-4} 
                                 &                               &                                   & 'I do not believe so.For I am sure that you and I should agree that good men are not averse to evil, and the desire of evil is often to have power over things not'                                                                         \\ \hline
\end{longtable}
\clearpage
\twocolumn

\end{document}